\documentclass{article}
\usepackage{preprint}
\usepackage{amsmath,graphicx}
\usepackage{float}
\usepackage{listings}
\usepackage{mathtools}
\usepackage{upgreek}
\usepackage{tabularx}
\usepackage{siunitx}
\usepackage{caption,setspace}
\usepackage{enumitem}
\usepackage{multirow}
\usepackage{pifont}
\usepackage{multicol}
\usepackage{booktabs}
\usepackage{subcaption}

\usepackage[numbers,sort&compress]{natbib}
\usepackage[toc,page]{appendix}
\usepackage[pagebackref,breaklinks,colorlinks,citecolor=cvprblue]{hyperref}
\usepackage[dvipsnames]{xcolor}

\renewcommand{\shorttitle}{ARTEMIS-DA}

\definecolor{cvprblue}{rgb}{0.21,0.49,0.74}
\usepackage[pagebackref,breaklinks,colorlinks,citecolor=cvprblue]{hyperref}

\title{ARTEMIS-DA: An Advanced Reasoning and Transformation Engine for Multi-Step Insight Synthesis in Data Analytics}
  
\author{Atin Sakkeer Hussain\\
{\tt\small atinsaki@gmail.com}
}

\begin{document}
\twocolumn[{
\maketitle
\begin{center}
    \captionsetup{type=figure}
    \includegraphics[width=\textwidth]{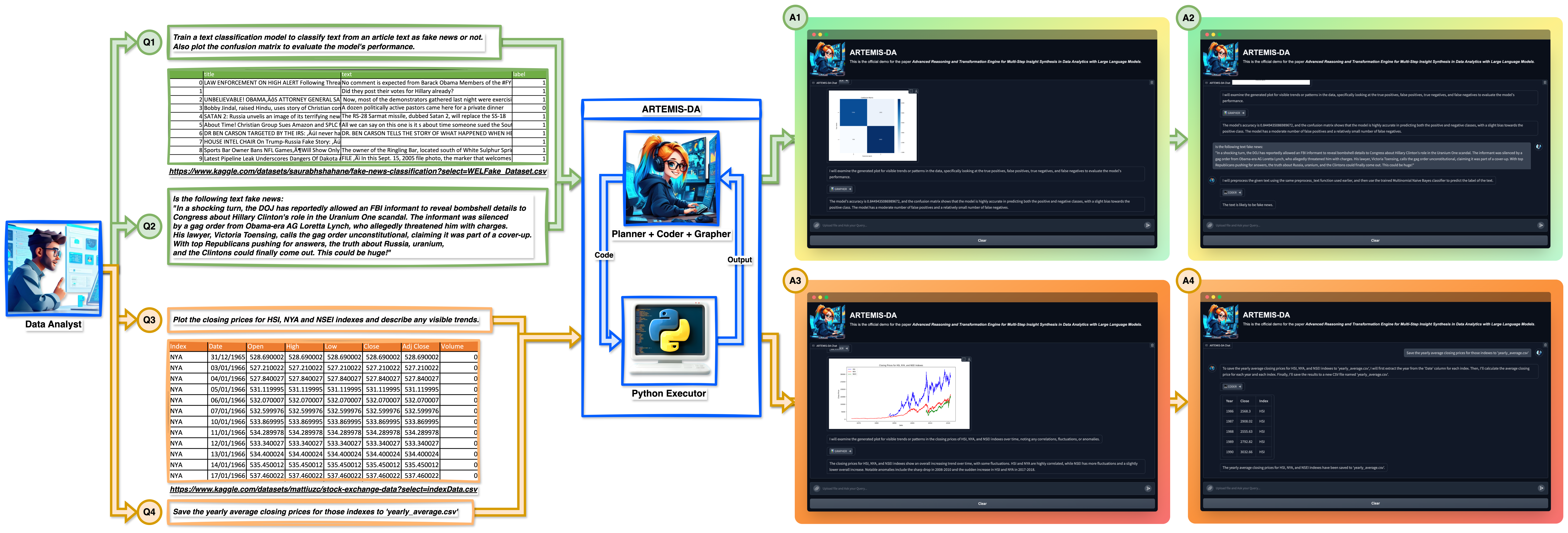}
    \captionof{figure}{ARTEMIS-DA showcasing \textit{\textbf{Q1}}. advanced reasoning for complex queries, \textit{\textbf{Q2}}. predictive modeling for text classification, \textit{\textbf{Q3}}. data visualization for insights, and \textit{\textbf{Q4}}. efficient data transformation for manipulating datasets.}
\end{center}
\vspace{4mm}
}]

\begin{abstract}

This paper presents the \textbf{Advanced Reasoning and Transformation Engine for Multi-Step Insight Synthesis in Data Analytics (ARTEMIS-DA)}, a novel framework designed to augment Large Language Models (LLMs) for solving complex, multi-step data analytics tasks. ARTEMIS-DA integrates three core components: the \textbf{Planner}, which dissects complex user queries into structured, sequential instructions encompassing data preprocessing, transformation, predictive modeling, and visualization; the \textbf{Coder}, which dynamically generates and executes Python code to implement these instructions; and the \textbf{Grapher}, which interprets generated visualizations to derive actionable insights. By orchestrating the collaboration between these components, ARTEMIS-DA effectively manages sophisticated analytical workflows involving advanced reasoning, multi-step transformations, and synthesis across diverse data modalities. The framework achieves state-of-the-art (SOTA) performance on benchmarks such as WikiTableQuestions and TabFact, demonstrating its ability to tackle intricate analytical tasks with precision and adaptability. By combining the reasoning capabilities of LLMs with automated code generation and execution and visual analysis, ARTEMIS-DA offers a robust, scalable solution for multi-step insight synthesis, addressing a wide range of challenges in data analytics.

\end{abstract}

\section{Introduction}

The advent of \textit{Large Language Models} (LLMs), such as GPT-3 \cite{brown2020languagemodelsfewshotlearners}, GPT-4 \cite{openai2024gpt4technicalreport}, and Llama 3 \cite{dubey2024llama3herdmodels}, has revolutionized the fields of artificial intelligence (AI) and natural language processing (NLP). These models have demonstrated remarkable capabilities in complex interpretation, reasoning, and generating human-like language, achieving success in diverse applications such as language translation, summarization, content generation, and question-answering. Their ability to process and generate coherent text has also made them powerful tools for tasks requiring nuanced understanding and reasoning. However, while LLMs have been extensively explored in these domains, their potential in transforming the field of data analytics remains underutilized. The inherent reasoning and code-generation capabilities of LLMs suggest immense promise for enabling non-technical users to interact with complex datasets using natural language, bridging the gap between advanced analytics and accessibility.

Recent efforts, including TableLLM \cite{liu2023rethinkingtabulardataunderstanding}, CABINET \cite{patnaik2024cabinetcontentrelevancebased}, and Chain-of-Table \cite{wang2024chainoftableevolvingtablesreasoning}, have begun to explore this intersection. These studies focus on table-based question answering and reasoning tasks, leveraging LLMs to interpret and respond to queries grounded in structured datasets. While these works demonstrate the potential of LLMs for data analytics, they primarily address single-step tasks, leaving multi-step analytical processes—such as complex data transformation, predictive modeling, and visualization—relatively unexplored. These tasks often require sequential reasoning, complex task decomposition, and precise execution across diverse operations, presenting challenges that current frameworks are not yet equipped to manage effectively.

To address this gap, we propose the \textbf{Advanced Reasoning and Transformation Engine for Multi-Step Insight Synthesis in Data Analytics (ARTEMIS-DA)}, a comprehensive framework explicitly designed to unlock the potential of LLMs for advanced multi-step data analytics tasks. ARTEMIS-DA introduces a tri-component architecture consisting of a \textbf{\textit{Planner}}, a \textbf{\textit{Coder}}, and a \textbf{\textit{Grapher}}, each playing a crucial, complementary role. The \textbf{\textit{Planner}} acts as the framework's coordinator, interpreting complex user queries and decomposing them into a sequence of structured instructions tailored to the dataset and analytical goals. These instructions encompass tasks such as data pre-processing, transformation, predictive analysis, and visualization. By leveraging the natural language understanding and reasoning capabilities of LLMs, the Planner ensures clarity and structure in task execution, addressing the intricacies of multi-step analytical workflows.

The \textbf{\textit{Coder}}, in turn, translates the Planner's instructions into executable Python code, dynamically generating and executing the necessary operations within a Python environment. Whether performing data transformations, training predictive models, or creating visualizations, the Coder bridges the gap between high-level task specifications and low-level computational execution. This dynamic interplay between the Planner and Coder components enables ARTEMIS-DA to handle analytical tasks autonomously, requiring minimal user intervention. 

The \textbf{\textit{Grapher}} is another key component of ARTEMIS-DA, analyzing the generated graphs and visualizations to extract valuable insights. By interpreting visual representations of data, the Grapher enables a deeper understanding of the underlying patterns and trends. The Grapher works in close coordination with the Planner and Coder, facilitating the seamless integration of data analysis, visualization, and insight extraction into a unified framework. Together, these components empower ARTEMIS-DA to deliver actionable insights, facilitating natural language interactions with complex datasets and enhancing accessibility for users without programming expertise.

ARTEMIS-DA's effectiveness is underscored by its State-of-the-Art (SOTA) performance on benchmarks such as WikiTableQuestions \cite{pasupat-liang-2015-compositional} and TabFact \cite{2019TabFactA}. These results highlight its ability to manage nuanced analytical tasks requiring advanced multi-step reasoning. Beyond answering structured dataset queries, ARTEMIS-DA demonstrates versatility in transforming datasets, visualizing results, and conducting predictive modeling, positioning it as a transformative tool in LLM-powered data analytics.

The remainder of this paper is structured as follows: Section 2 reviews related work on LLM applications in data analytics and automated task decomposition. Section 3 presents the architecture of ARTEMIS-DA, focusing on its components and their respective roles. Section 4 evaluates the framework’s performance on established benchmarks, while Section 5 demonstrates ARTEMIS-DA’s capabilities in generating visualizations and predictive modeling. Finally, Section 6 concludes with insights into future directions for advancing LLM-powered data analytics and further enhancing LLMs for other complex tasks that require multi-step reasoning.

Through the development of ARTEMIS-DA, we contribute a significant advancement to the emerging field of LLM-driven data analytics, offering a fully integrated end-to-end system capable of managing complex, multi-step analytical queries with minimal user intervention. This framework aims to redefine the accessibility, efficiency, and scope of data analytics, establishing a new paradigm for LLM-assisted insight synthesis.
\section{Related Works}
\label{sec:related}
In recent years, Large Language Models (LLMs) have demonstrated promise in addressing complex tasks in natural language processing and reasoning. However, when applied to structured data, such as large tables, unique challenges arise, requiring specialized frameworks and adaptations. Recent studies have made significant strides in enhancing LLMs' reasoning capabilities with tabular data, providing the foundation for the ARTEMIS-DA framework proposed in this paper.

\textbf{Tabular Reasoning with Pre-trained Language Models.} Traditional approaches with pre-trained language models such as TaBERT\cite{yin20acl}, TAPAS\cite{Herzig_2020}, TAPEX\cite{liu2022tapextablepretraininglearning}, ReasTAP\cite{zhao-etal-2022-reastap}, and PASTA\cite{gu2022pastatableoperationsawarefact} were developed to combine free-form questions with structured tabular data. These models achieved moderate success by integrating table-based and textual training, enhancing tabular reasoning capabilities. However, their ability to generalize under table perturbations remains limited\cite{zhao2023robutsystematicstudytable, chang2023drspiderdiagnosticevaluationbenchmark}. Solutions such as LETA\cite{zhao2023robutsystematicstudytable} and LATTICE\cite{wang2022robust} addressed these limitations using data augmentation and order-invariant attention. However, they require white-box access to models, making them incompatible with SOTA black-box LLMs.

\textbf{Table-Specific Architectures for LLMs.} Table-specific models further refined the use of structured tabular data, emphasizing row and column positioning such as TableFormer\cite{yang-etal-2022-tableformer} which introduced positional embeddings to capture table structure, mitigating the impact of structural perturbations. Despite these advances, frameworks like StructGPT\cite{jiang2023structgptgeneralframeworklarge} highlighted that generic LLMs still struggle with structured data unless enhanced with explicit symbolic reasoning. Recently proposed frameworks such as AutoGPT\cite{Significant_Gravitas_AutoGPT} and DataCopilot\cite{zhang2023data} began addressing table-specific challenges by incorporating advanced reasoning techniques. However, their performance remains constrained across diverse scenarios due to their reliance on generic programming for task execution.

\textbf{Noise Reduction in Table-Based Question Answering.} Handling noise in large, complex tables is another active research area. CABINET\cite{patnaik2024cabinetcontentrelevancebased} introduced a Content Relevance-Based Noise Reduction strategy, significantly improving LLM performance by employing an \textit{Unsupervised Relevance Scorer} (URS). By filtering irrelevant information and focusing on content relevant to user queries, CABINET achieved notable accuracy improvements on datasets like WikiTableQuestions\cite{pasupat-liang-2015-compositional} and FeTaQA\cite{10.1162/tacl_a_00446}, underscoring the importance of noise reduction in reliable table-based reasoning.

\textbf{Few-Shot and Zero-Shot Learning for Tabular Reasoning.} Few-shot and zero-shot learning methods have shown considerable promise in tabular understanding tasks. Chain-of-Thought (CoT) prompting\cite{wei2023chainofthoughtpromptingelicitsreasoning} improved LLMs' sequential reasoning capabilities. Building on this, studies such as\cite{cheng2023bindinglanguagemodelssymbolic, ye2023largelanguagemodelsversatile} integrated symbolic reasoning into CoT frameworks, enhancing query decomposition and understanding. However, these techniques are not tailored to tabular structures, leading to performance gaps. Chain-of-Table\cite{wang2024chainoftableevolvingtablesreasoning} addressed these limitations by introducing an iterative approach to transforming table contexts, enabling more effective reasoning over structured data and achieving state-of-the-art results on benchmarks such as WikiTableQuestions\cite{pasupat-liang-2015-compositional} and TabFact\cite{2019TabFactA}.

\textbf{Programmatic Approaches to Tabular Question Answering.} Leveraging programmatic techniques has significantly advanced the field of table-based question answering by enabling models to interpret and process structured data more effectively. Text-to-SQL models such as TAPEX\cite{liu2022tapextablepretraininglearning} and OmniTab\cite{jiang-etal-2022-omnitab} trained large language models (LLMs) to translate natural language questions into SQL operations, demonstrating the potential for automated interaction with tabular datasets. Despite their promise, these models struggled with noisy and large tables due to limited query comprehension. Programmatically enhanced solutions, such as Binder\cite{cheng2023bindinglanguagemodelssymbolic} and LEVER\cite{ni2023lever}, improved performance by generating and verifying SQL or Python code. However, their reliance on single-pass code generation restricted their adaptability to queries requiring dynamic reasoning.

Building on these advancements and addressing the limitations of previous models, we introduce the \textbf{Advanced Reasoning and Transformation Engine for Multi-Step Insight Synthesis in Data Analytics (ARTEMIS-DA)}. Unlike prior models, ARTEMIS-DA features a tri-component architecture consisting of a \textbf{\textit{Planner}}, a \textbf{\textit{Coder}}, and a \textbf{\textit{Grapher}}, which collaboratively address complex, multi-step analytical queries. The \textbf{\textit{Planner}} generates dynamic task sequences tailored to specific datasets and queries, encompassing data transformation, predictive analysis, and visualization. The \textbf{\textit{Coder}} translates these sequences into Python code and executes them in real time. Finally, the \textbf{\textit{Grapher}} extracts actionable insights from the visualizations produced by the Coder, enhancing result interpretability and ensuring the seamless integration of visual insights into user workflows.

This seamless integration of the Planner, Coder, and Grapher components enables ARTEMIS-DA to tackle intricate tasks requiring sequential reasoning, synthesis, and visualization, achieving state-of-the-art performance on benchmarks such as WikiTableQuestions\cite{pasupat-liang-2015-compositional} and TabFact\cite{2019TabFactA}. ARTEMIS-DA extends LLM functionality in data analytics, providing a robust, automated solution that empowers technical and non-technical users to interact with complex datasets through natural language.

\section{ARTEMIS-DA Framework}

\begin{figure}[h]
\centering
\includegraphics[width=\linewidth]{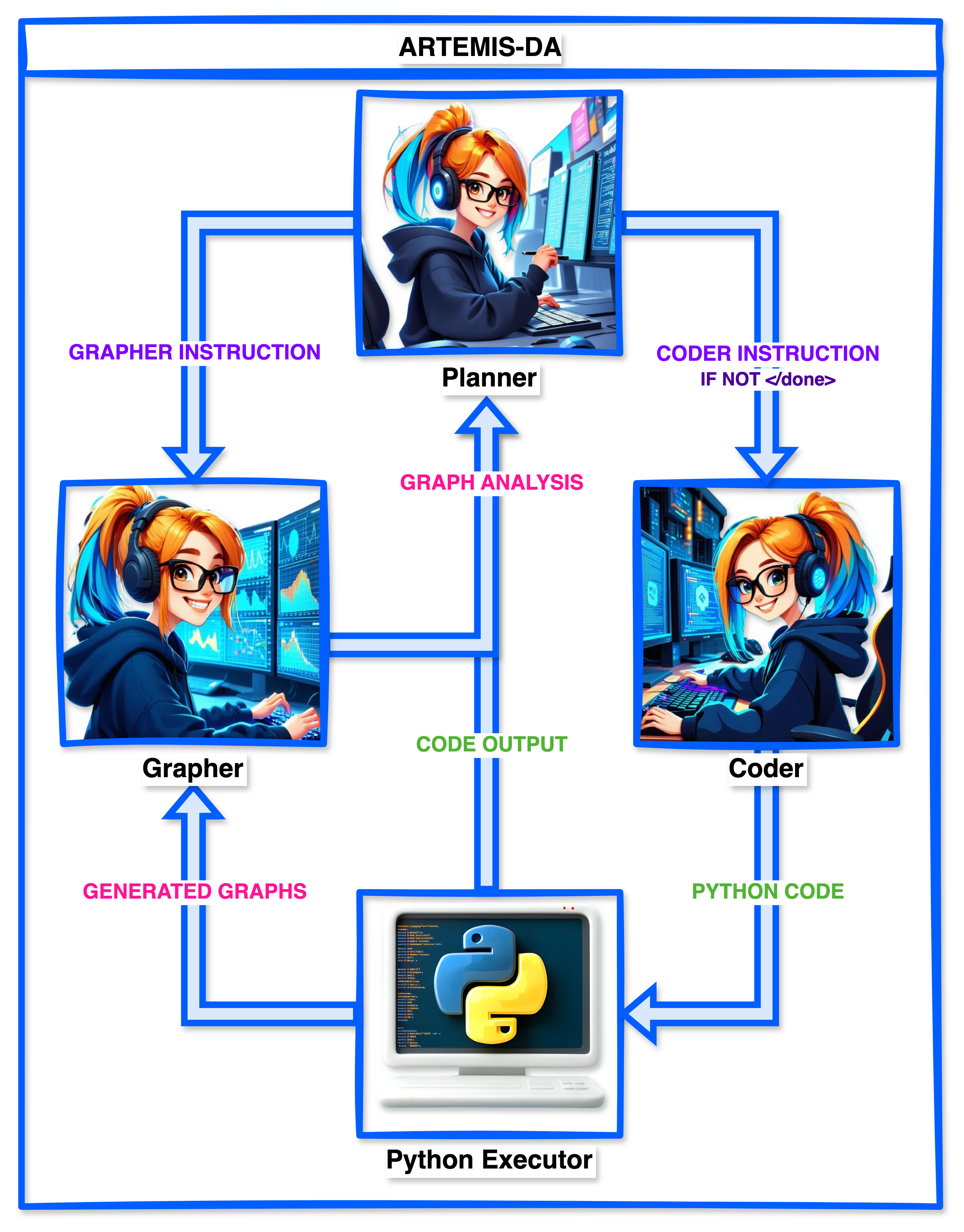}
\caption{ARTEMIS-DA Framework}
\label{fig:model_architecture}
\end{figure}

The \textbf{Advanced Reasoning and Transformation Engine for Multi-Step Insight Synthesis in Data Analytics (ARTEMIS-DA)} is designed to address the challenges of complex data analytics by combining advanced reasoning capabilities with dynamic, real-time code generation, execution and visual analysis. The framework, shown in Figure \ref{fig:model_architecture}, consists of three core components: the \textbf{\textit{Planner}}, the \textbf{\textit{Coder}} and  the \textbf{\textit{Grapher}}. Working in unison, these components decompose complex queries into sequential tasks, automatically generate and execute the required code for each task, and synthesize insights based on generated graphs with minimal user intervention. 

For the experiments in this paper, all three components utilize the LLaMA 3 70B model\cite{dubey2024llama3herdmodels}, which demonstrated superior performance across benchmarks. The Grapher component also employs the LLaMA 3.2 Vision 90B model for understanding generated graphs. However, the framework is model-agnostic and can be adapted to work with other state-of-the-art large language models (LLMs), such as GPT-4\cite{openai2024gpt4technicalreport} or Mixtral-8x7B\cite{jiang2024mixtralexperts}.

The following sections provide an in-depth exploration of the Planner, Coder and Grapher components, detailing their roles and interactions. Together, they exemplify ARTEMIS-DA’s ability to streamline end-to-end data analytics workflows effectively and efficiently.

\begin{figure*}[ht]
\centering
\includegraphics[width=\textwidth]{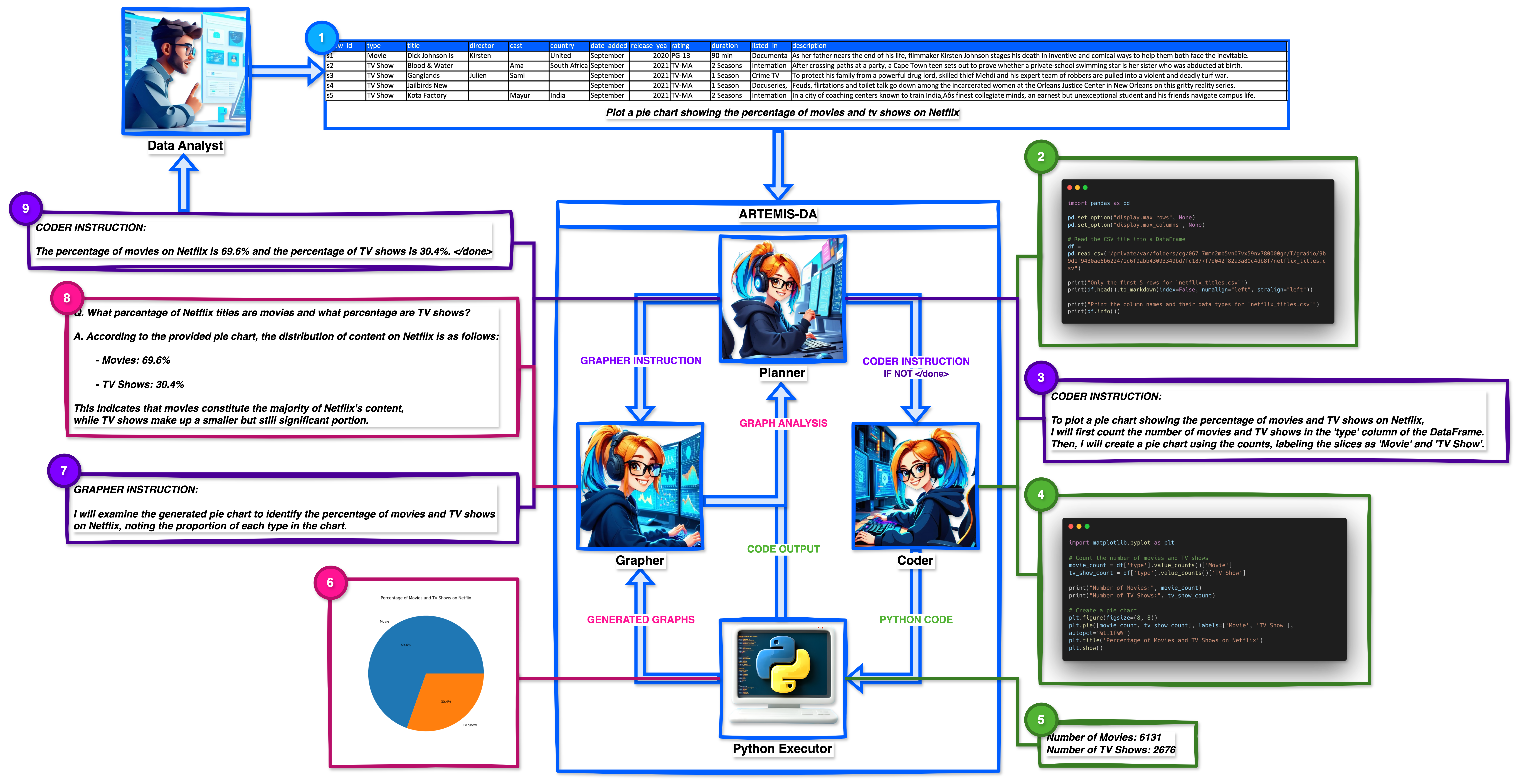}
\caption{Workflow of the \textbf{ARTEMIS-DA} framework showcasing its multi-component collaboration.}
\label{fig:model_workflow}
\end{figure*}

\subsection{Planner Component}

The \textbf{\textit{Planner}} serves as the central reasoning and task-decomposition unit within the ARTEMIS-DA framework, converting user queries into structured sequences of tasks. Its primary responsibilities include parsing user inputs, organizing them into a logical workflow, and interpreting the outputs of executed code and generated visual insights to guide subsequent steps in the analytical process. The Planner is adept at managing complex, multi-faceted queries that require diverse operations, such as data transformation, predictive modeling, and visualization. By leveraging the advanced reasoning capabilities of large language models (LLMs), it decomposes intricate requirements into optimized task sequences aligned with the input data, prior outputs, and specific details of the user’s query, ensuring clarity for seamless execution by the Coder and Grapher as required.

For instance, when tasked with analyzing sales patterns and forecasting future trends, the Planner systematically outlines essential steps, such as data pre-processing, feature engineering, model training, evaluation and forecasting, organizing them into a coherent workflow. This process is dynamically tailored to the dataset’s properties and the nuances of the query, enabling an adaptive approach to multi-step analytical tasks. Each task specification is then relayed to the Coder or Grapher in sequence, ensuring a smooth and collaborative workflow across the components of the ARTEMIS-DA framework.

\subsection{Coder Component}

Following the Planner’s generation of a structured task sequence, the \textbf{\textit{Coder}} translates these tasks into executable Python code. It processes instructions for each step—such as loading data, creating visualizations, or training predictive models—producing contextually relevant and functionally precise code tailored to the task's requirements. Operating in real time within a Python environment, the Coder executes each code snippet, generating intermediate outputs that are fed back to the Planner for further analysis and informed decision-making.

The Coder's capabilities encompass a broad range of analytical operations, from basic data cleaning and summarization to advanced predictive modeling and visualization. For instance, when the user requests predictive analysis on time-series data, the Planner decomposes the query into multiple simple steps such as splitting the data, training a suitable model, and visualizing the results. The Coder generates and executes code for each step, with the Planner overseeing all steps in the process. This iterative exchange between the Planner and Coder ensures accurate execution of each analytical step, maintaining a continuous feedback loop until the analysis is fully completed.

\subsection{Grapher Component}

The \textbf{\textit{Grapher}} serves as a critical component for deriving actionable insights from visual data, responding to instructions generated by the \textbf{\textit{Planner}}. Upon receiving a directive to analyze a generated graph from the Planner, the Grapher processes the visual output and provides insights in a structured question-and-answer format. Its analytical capabilities encompass a broad spectrum, ranging from basic data extraction from plots to advanced interpretation and trend analysis of complex plots.

For instance, if the Planner requests an analysis of time-series data, and the Coder generates a corresponding line plot, the Grapher can identify trends, detect anomalies, and highlight significant observations. This feedback enables the Planner to refine its understanding and produce more nuanced insights. Additionally, the Grapher supports a wide variety of graph types, including bar charts, pie charts, scatter plots, and heatmaps, among others, ensuring versatility across diverse data visualization needs. By seamlessly integrating graph interpretation into the workflow, the Grapher enhances the framework's capacity to provide meaningful, data-driven conclusions.

\begin{table*}[ht]
\renewcommand{\arraystretch}{1.25}
\centering
\begin{tabular}{l | c | c | c | c}
\hline \hline
\multirow{2}{*}{\textbf{Model}}                                    & \textbf{WikiTableQuestions}                               & \textbf{TabFact}                                        & \multicolumn{2}{c}{\textbf{FeTaQA}}  \\ \cline{2-5}
                                                                   & \textbf{Accuracy}                                         & \textbf{Accuracy}                                       & \textbf{S-BLEU}                                         & \textbf{BLEU}                                             \\ \hline \hline
\textbf{With Fine-tuned / Trained}                    &                                                           &                                                         &                                                         &                                                           \\
ReasTAP-Large\cite{zhao-etal-2022-reastap}                         & 58.7                                                      & 86.2                                                    & -                                                       & -                                                         \\ 
OmniTab-Large\cite{jiang-etal-2022-omnitab}                        & 63.3                                                      & -                                                       & 34.9                                                    & -                                                         \\ 
LEVER\cite{ni2023lever}                                            & 65.8                                                      & -                                                       & -                                                       & -                                                         \\
CABINET\cite{patnaik2024cabinetcontentrelevancebased}              & 69.1                                                      & -                                                       & \underline{40.5}                                        & -                                                         \\ \hline \hline
\textbf{Without Fine-tuning/Training}                  &                                                           &                                                         &                                                         &                                                           \\
Binder\cite{cheng2023bindinglanguagemodelssymbolic}                & 64.6                                                      & 86.0                                                    & -                                                       & -                                                         \\ 
DATER\cite{ye2023largelanguagemodelsversatile}                     & 65.9                                                      & 87.4                                                    & 30.9                                                    & 29.5                                                      \\ 
Tab-PoT\cite{Xiao_2024}                                            & 66.8                                                      & 85.8                                                    & -                                                       & -                                                         \\ 
Chain-of-Table\cite{wang2024chainoftableevolvingtablesreasoning}   & 67.3                                                      & 86.6                                                    & -                                                       & \underline{32.6}                                          \\ 
SynTQA (RF)\cite{zhang2024syntqasynergistictablebasedquestion}     & 71.6                                                      & -                                                       & -                                                       & -                                                         \\
Mix-SC\cite{liu2023rethinkingtabulardataunderstanding}             & 73.6                                                      & \underline{88.5}                                        & -                                                       & -                                                         \\ 
SynTQA (GPT)\cite{zhang2024syntqasynergistictablebasedquestion}    & \underline{74.4}                                          & -                                                       & -                                                       & -                                                         \\ \hline \hline
\textbf{ARTEMIS-DA (Ours)}                                            & \textbf{80.8 {\footnotesize{\color{OliveGreen}(+6.4)}}}  & \textbf{93.1 {\footnotesize{\color{OliveGreen}(+4.6)}}} & \textbf{62.7 {\footnotesize{\color{OliveGreen}(+22.2)}}} & \textbf{46.4 {\footnotesize{\color{OliveGreen}(+13.8)}}} \\
\hline \hline
\end{tabular}
\caption{Performance comparison for models on WikiTableQuestions, TabFact and FeTaQA datasets.}
\label{tab:model_performance}
\end{table*}

\subsection{Workflow and Interaction between Components}

The ARTEMIS-DA framework employs a systematic workflow to process user queries with precision and efficiency. This process, illustrated in Figure \ref{fig:model_workflow}, highlights the seamless interaction between the Planner, Coder, and Grapher components as outlined below:

\begin{enumerate}[leftmargin=*]
    \item \textbf{Input:} The workflow begins when the user submits a dataset and a natural language query describing their analytical objectives. The Coder starts by generating the Python code to load the dataset and display the column types and the first five rows of the dataset.
    
    \item \textbf{Decomposition:} The Planner analyzes the query and decomposes it into a structured sequence of tasks, leveraging outputs from \texttt{df.info()} and \texttt{df.head()} to extract actionable context. For instance, a query to compare media categories might involve counting the number of TV shows and movies, creating a pie chart for visualization, and analyzing the generated chart for insights. Each task is methodically assigned to the Coder or Grapher as appropriate.
    
    \item \textbf{Execution:} The Coder translates the Planner’s instructions into executable Python code. Tasks are executed sequentially, producing intermediate outputs. In the sample query, the Coder generates the code to count the specified categories, generate the pie chart, and prepare the visual output for analysis.
    
    \item \textbf{Analysis:} The Grapher processes the Planner’s instructions to derive insights from generated visuals. In the sample query, the Grapher analyzes the pie chart to calculate and report the proportions of TV shows and movies, presenting insights in a structured format.
    
    \item \textbf{Feedback Loop:} Intermediate outputs, such as computed values, visualizations, and insights, are returned to the Planner. The Planner evaluates these results to determine if additional steps are necessary. If further actions are required, new instructions are dynamically generated for the Coder or Grapher, maintaining a responsive feedback loop to achieve the task.
    
    \item \textbf{Finalization:} Once all tasks are completed, the Planner aggregates the results, refines the insights, and compiles the final output. The final result, enhanced with additional insights if necessary, is presented to the user along with a \texttt{</done>} tag, indicating the successful conclusion of the workflow.
\end{enumerate}

\subsection{Advantages of the ARTEMIS-DA Framework}

The ARTEMIS-DA framework represents a significant advancement in data analytics by integrating sophisticated natural language understanding with precise computational execution and visual insight synthesis. Its tri-component architecture—comprising the Planner, Coder, and Grapher—enables the framework to efficiently handle complex, multi-step analytical tasks, all while providing an intuitive interface suitable for users across various levels of technical expertise.

ARTEMIS-DA achieves state-of-the-art performance on challenging datasets such as WikiTableQuestions\cite{pasupat-liang-2015-compositional}, TabFact\cite{2019TabFactA}, and FeTaQA\cite{10.1162/tacl_a_00446}, demonstrating its versatility in extracting comprehensive, data-driven insights. The framework’s ability to dynamically interpret user queries, generate and execute Python code in real-time, and analyze generated visuals further enhances its adaptability and practical value. These features collectively position ARTEMIS-DA as a cutting-edge solution for addressing the modern challenges of data analytics with precision, efficiency, and user-centered design.

\begin{table*}[ht]
\renewcommand{\arraystretch}{1.25}
\centering
\begin{tabular}{l | c | c | c | c}
\hline \hline
\multirow{2}{*}{\textbf{Model}}  & \textbf{WikiTableQuestions}                                & \textbf{TabFact}                                                  & \multicolumn{2}{c}{\textbf{FeTaQA}}  \\ \cline{2-5}
                                 & \textbf{Accuracy}                                          & \textbf{Accuracy}                                                 & \textbf{S-BLEU}                                          & \textbf{BLEU}                                            \\ \hline \hline
Best SOTA Model                  & 74.4 \cite{zhang2024syntqasynergistictablebasedquestion}   & \underline{88.5} \cite{liu2023rethinkingtabulardataunderstanding} & 40.5 \cite{patnaik2024cabinetcontentrelevancebased}      & 32.6 \cite{wang2024chainoftableevolvingtablesreasoning}  \\
LLaMA 3 70B                      & 72.1                                                       & 85.1                                                              & 11.0                                                     & 20.4                                                     \\ 
Single Step ARTEMIS-DA           & \underline{76.6}                                           & 87.6                                                              & \underline{40.8}                                         & \underline{32.8}                                         \\ \hline \hline
\textbf{ARTEMIS-DA (Ours)}       & \textbf{80.8 {\footnotesize{\color{OliveGreen}(+4.2)}}}    & \textbf{93.1 {\footnotesize{\color{OliveGreen}(+4.6)}}}           & \textbf{62.7 {\footnotesize{\color{OliveGreen}(+21.9)}}} & \textbf{46.4 {\footnotesize{\color{OliveGreen}(+13.6)}}} \\
\hline \hline
\end{tabular}

\caption{Ablation study of ARTEMIS-DA on WikiTableQuestions, TabFact and FeTaQA datasets.}
\label{tab:model_ablation}
\end{table*}

\section{Experiments and Evaluation}
\label{sec:experiments}

This section provides an overview of the benchmark datasets utilized for evaluation, the metrics employed to compare the performance of the models, the results achieved by the ARTEMIS-DA framework, and a comprehensive analysis of these results.

\subsection{Datasets}
The ARTEMIS-DA framework is evaluated on three table-based reasoning datasets: WikiTableQuestions \cite{pasupat-liang-2015-compositional}, TabFact \cite{2019TabFactA}, and FeTaQA \cite{10.1162/tacl_a_00446}. We evaluate ARTEMIS-DA exclusively on the test sets of these datasets without any training or fine-tuning on the training sets. The details of each dataset are as follows:

\begin{itemize}[leftmargin=*]
    \item \textbf{TabFact \cite{2019TabFactA}:} This dataset serves as a benchmark for table-based fact verification, comprising statements created by crowd workers based on Wikipedia tables. For example, a statement such as “the industrial and commercial panel has four more members than the cultural and educational panel” must be classified as “True” or “False” based on the corresponding table. We report the accuracy on the test-small set, which consists of 2,024 statements and 298 tables.

    \item \textbf{WikiTableQuestions \cite{pasupat-liang-2015-compositional}:} This dataset includes complex questions generated by crowd workers that are based on Wikipedia tables. The questions require various complex operations such as comparison, aggregation, and arithmetic, necessitating compositional reasoning across multiple entries in the given table. We utilize the standard test set, containing 4,344 samples.

    \item \textbf{FeTaQA \cite{10.1162/tacl_a_00446}:} This dataset consists of free-form tables questions that demand advanced reasoning. Most questions are based on information extracted from discontinuous sections of the table. We evaluate ARTEMIS-DA on the test set, which includes 2,003 samples.
\end{itemize}

\subsection{Evaluation Metrics}
To assess the performance of ARTEMIS-DA across the datasets, we employ a range of metrics tailored to the specific tasks. For the TabFact \cite{2019TabFactA} dataset, which focuses on table-based fact verification, we utilize binary classification accuracy to evaluate the correctness of statements in relation to the provided tables. In the case of WikiTableQuestions \cite{pasupat-liang-2015-compositional}, we measure denotation accuracy to determine whether the predicted answers align with the ground truth, leveraging the LLaMA 3 70B\cite{dubey2024llama3herdmodels} model for verification. Since FeTaQA \cite{10.1162/tacl_a_00446} aims to generate comprehensive, long-form answers rather than short phrases, we adopt both \textbf{BLEU} \cite{bleu} and \textbf{SacreBLEU} \cite{sacre-bleu} as our evaluation metrics. BLEU, a widely used metric in machine translation, assesses the quality of generated text by comparing it to high-quality reference translations, producing scores that range from 0 to 1, with higher values indicating greater similarity. SacreBLEU addresses inconsistencies commonly found in BLEU score reporting by providing a standardized framework for computation, ensuring that results are shareable, comparable, and reproducible across studies. The use of two distinct metrics for the FeTaQA \cite{10.1162/tacl_a_00446} dataset is warranted due to the variation in metrics employed in different studies. By leveraging these metrics, we can effectively evaluate ARTEMIS-DA's capabilities in table-based reasoning tasks and benchmark its performance against existing models.

\begin{figure*}[ht]
\centering
\includegraphics[width=\textwidth]{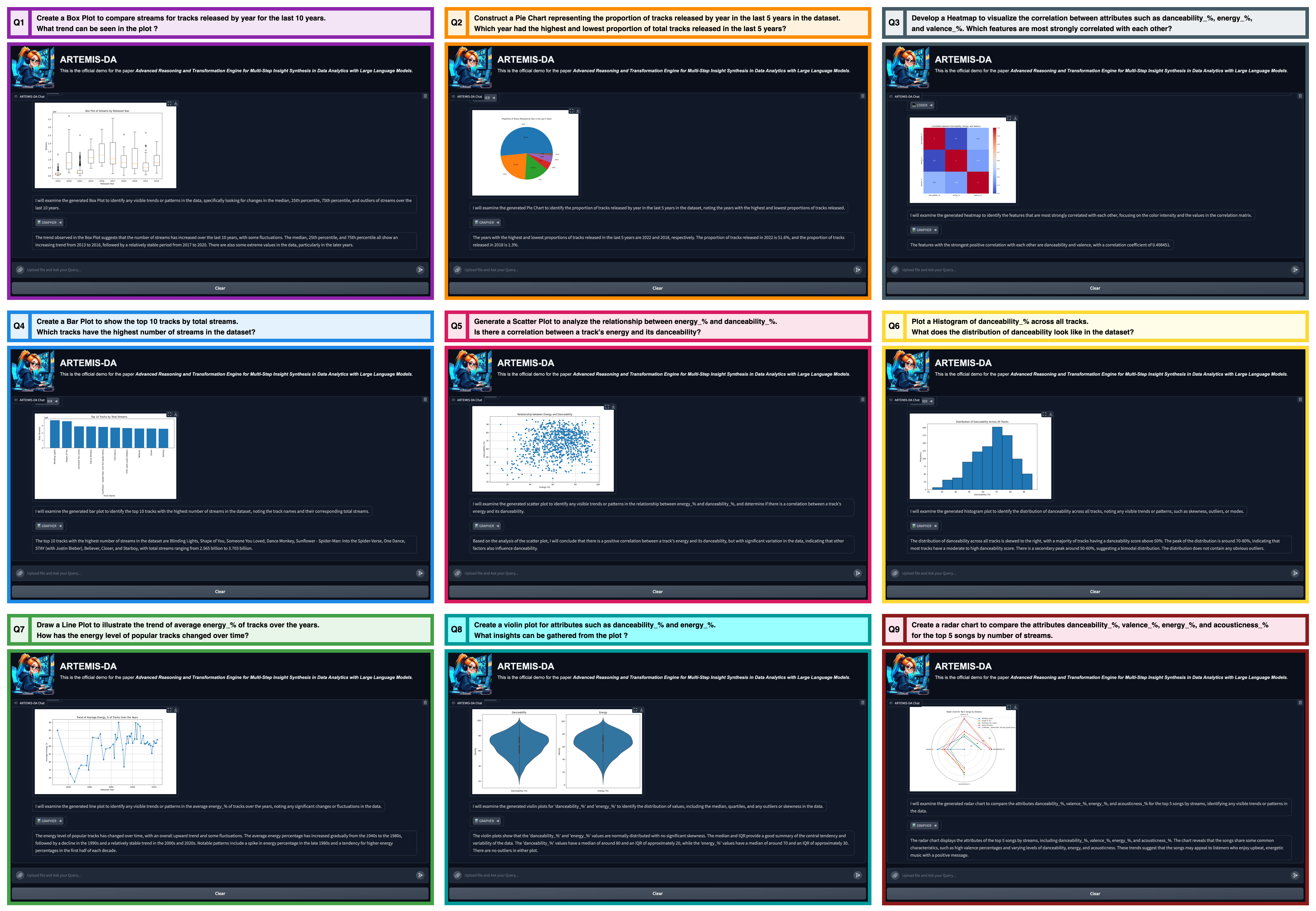}
\caption{ARTEMIS-DA's visualizations using the \href{https://www.kaggle.com/datasets/abdulszz/spotify-most-streamed-songs?select=Spotify+Most+Streamed+Songs.csv}{\underline{Spotify Most Streamed Songs}} dataset.}
\label{fig:plot_vis}
\end{figure*}

\subsection{Results}

Table \ref{tab:model_performance} highlights ARTEMIS-DA’s state-of-the-art performance across the WikiTableQuestions\cite{pasupat-liang-2015-compositional}, TabFact\cite{2019TabFactA}, and FeTaQA\cite{10.1162/tacl_a_00446} datasets. These results highlight the framework's adaptability and effectiveness in understanding, processing, and reasoning over structured tabular data across diverse domains and use cases.

In the WikiTableQuestions\cite{pasupat-liang-2015-compositional} dataset, ARTEMIS-DA achieves an accuracy of \textbf{80.8\%}, outperforming previous models such as SynTQA (GPT)\cite{zhang2024syntqasynergistictablebasedquestion}, CABINET\cite{patnaik2024cabinetcontentrelevancebased}, and LEVER\cite{ni2023lever}. This significant improvement of \textbf{+6.4\%} over the prior best underscores ARTEMIS-DA's superior compositional reasoning abilities, enabling it to effectively address complex table-based questions that require multi-step logical reasoning. 

Similarly, for the TabFact\cite{2019TabFactA} dataset, ARTEMIS-DA achieves an accuracy of \textbf{93.1\%}, surpassing Mix-SC\cite{liu2023rethinkingtabulardataunderstanding} by \textbf{+4.6\%}. This result demonstrates ARTEMIS-DA’s efficiency in verifying factual statements against tabular data with high precision, showcasing its strength in fact-checking tasks across diverse data representations.

For the FeTaQA\cite{10.1162/tacl_a_00446} dataset, ARTEMIS-DA delivers exceptional results in generating high-quality long-form answers. It achieves S-BLEU\cite{sacre-bleu} and BLEU\cite{bleu} scores of \textbf{62.7} and \textbf{46.4}, respectively, representing a substantial improvement over previous models, including CABINET\cite{patnaik2024cabinetcontentrelevancebased} and Chain-of-Table\cite{wang2024chainoftableevolvingtablesreasoning}, with gains of \textbf{+22.2} and \textbf{+13.8}, respectively. These results highlight ARTEMIS-DA's ability to extract, synthesize, and integrate discontinuous information from tables, providing coherent and contextually rich answers.

Overall, ARTEMIS-DA establishes itself as a highly versatile and effective solution for table-based reasoning and computational tasks. Leveraging the LLaMA 3 70B\cite{dubey2024llama3herdmodels} and LLaMA 3.2 Vision 90B models, ARTEMIS-DA consistently delivers state-of-the-art results across multiple datasets, further cementing its role as a transformative tool for compositional reasoning, fact verification, and long-form answer generation in diverse domains.

\subsection{Ablation Study}

To thoroughly assess the necessity of the multi-step design in the ARTEMIS-DA framework for addressing complex data analytics tasks, we conduct a detailed ablation study. The Single-Step ARTEMIS-DA variant eliminates the multi-step reasoning capability and instead processes each query in a single step. By including this simplified variant, we aim to isolate and highlight the specific contributions of the multi-step reasoning approach employed by ARTEMIS-DA when applied to a diverse set of benchmark datasets.

As shown in Table \ref{tab:model_ablation}, Single-Step ARTEMIS-DA achieves 76.6\% accuracy on WikiTableQuestions, 87.6\% on TabFact, and achieves 40.8 S-BLEU and 32.8 BLEU on FeTaQA, surpassing LLaMA 3 70B (72.1\%, 85.1\%, 11.0 S-BLEU, and 20.4 BLEU, respectively) but falling short of ARTEMIS-DA. The complete model outperforms both, achieving \textbf{80.8\%} accuracy on WikiTableQuestions (\textbf{4.2\%} gain) and \textbf{93.1\%} on TabFact (\textbf{4.6\%} gain), exceeding the SOTA benchmarks of 74.4\% and 88.5\% respectively. On FeTaQA, ARTEMIS-DA achieves \textbf{62.7} S-BLEU and \textbf{46.4} BLEU, representing significant improvements of \textbf{+21.9} S-BLEU and \textbf{+13.6} BLEU over Single-Step ARTEMIS-DA. These results confirm the value of ARTEMIS-DA's multi-step approach in handling complex queries, essential for accurate data analysis with LLMs.
\begin{figure*}[ht]
\centering
\includegraphics[width=\textwidth]{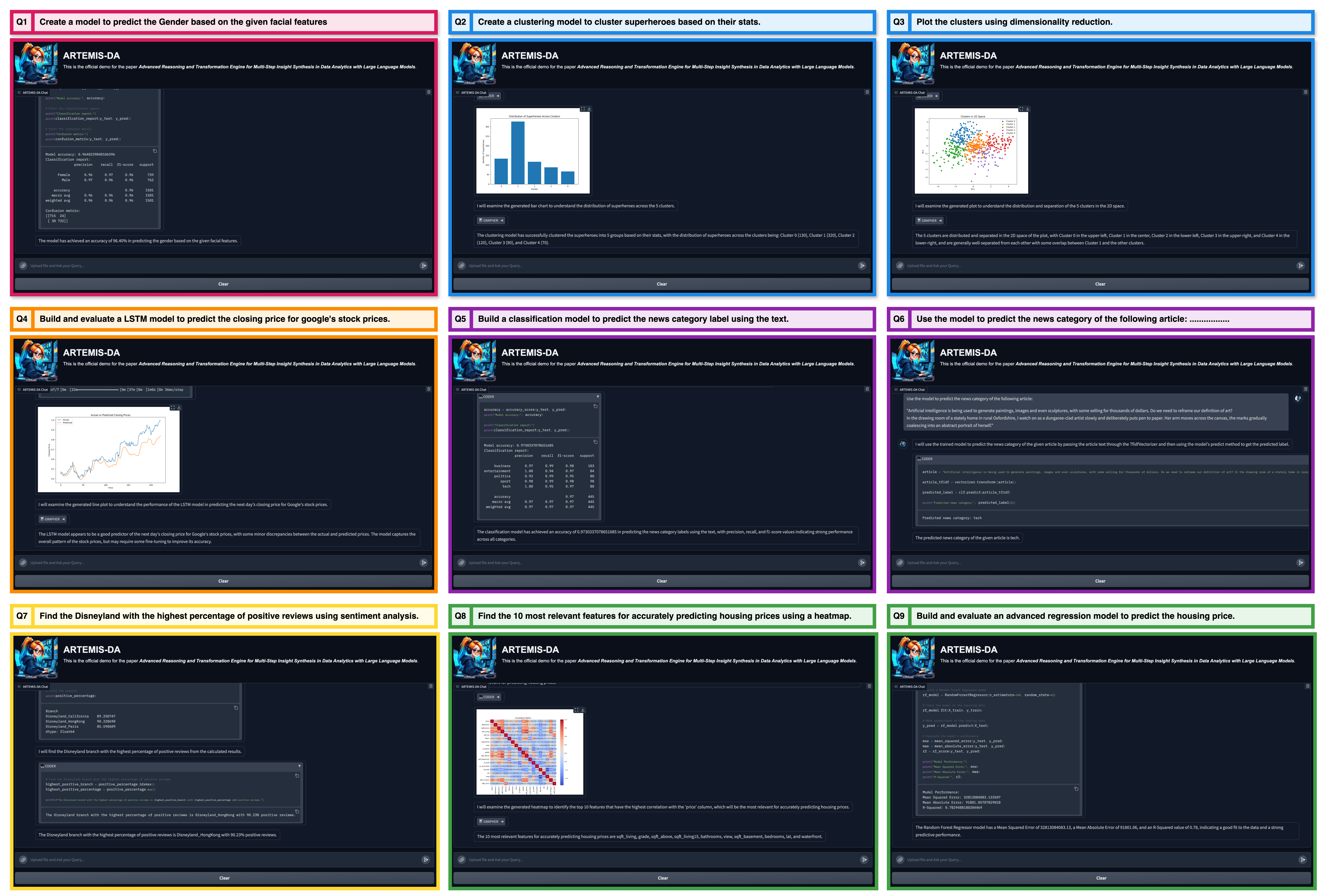}
\caption{ARTEMIS-DA's predictive modeling capabilities demonstrated through classification for numerical features and text datasets, time series forecasting on stock exchange data and clustering of superhero stats.}
\label{fig:combined_pred_model}
\end{figure*}

\section{ARTEMIS Capability Demonstration}

This section presents ARTEMIS-DA's capabilities in plot visualization and predictive modeling across a range of datasets, highlighting its versatility and robustness.

\subsection{Plot Visualization}

Figure \ref{fig:plot_vis} presents a diverse collection of visualizations generated by ARTEMIS-DA, demonstrating its analytical capabilities using the \href{https://www.kaggle.com/datasets/abdulszz/spotify-most-streamed-songs?select=Spotify+Most+Streamed+Songs.csv}{\underline{Spotify Most Streamed Songs}} dataset from Kaggle. These visualizations address a range of analytical questions, showcasing ARTEMIS-DA’s ability to effectively generate, interpret, and extract insights from various plot types. For instance, the box plot compares streams for tracks released in the last 10 years, identifying streaming trends over time. The pie chart analyzes the proportion of tracks released by year over the past five years, highlighting production patterns and identifying peak and low-release years. A heatmap reveals correlations among attributes like danceability\%, energy\%, and valence\%, uncovering relationships between musical features. Additionally, the bar plot highlights the top 10 tracks by total streams, offering a view of popular listener preferences, while the scatter plot explores potential correlations between energy\% and danceability\%. The histogram details the distribution of danceability\% across all tracks, providing insights into its variability. A line plot illustrates the trend of average energy\% of tracks over the years, shedding light on changes in the energy levels over time. The violin plot compares the distributions of danceability\% and energy\%, providing a nuanced understanding of their variability. Lastly, the radar chart contrasts attributes such as danceability\%, valence\%, energy\%, and acousticness\% for the top 5 tracks by total streams, offering a comparison of these popular tracks. Together, these plots underscore ARTEMIS-DA’s versatility and effectiveness in generating insightful, data-driven visualizations from complex datasets.

\subsection{Predictive Modeling}

Figure~\ref{fig:combined_pred_model} illustrates ARTEMIS-DA’s predictive modeling capabilities, emphasizing its effectiveness across diverse datasets and analytical tasks. These examples demonstrate the framework's ability to adapt seamlessly to various problem domains, showcasing ARTEMIS-DA as a robust solution for a wide range of predictive applications.

\begin{itemize}[leftmargin=*]
    \item \textbf{Classification of Gender Data:} The top-left sub-figure highlights ARTEMIS-DA’s ability to handle binary classification tasks using the \href{https://www.kaggle.com/datasets/elakiricoder/gender-classification-dataset?select=gender_classification_v7.csv}{\underline{Gender Classification}} dataset. By building a model to predict gender based on input features, ARTEMIS-DA achieves an accuracy of 96.7\%, demonstrating its proficiency in solving classification problems and its applicability to tasks such as customer profiling and demographics analysis.

    \item \textbf{Clustering of Superhero Powers:} The top-right sub-figures demonstrate ARTEMIS-DA’s clustering capabilities using the \href{https://www.kaggle.com/datasets/shreyasur965/super-heroes-dataset?select=superheroes_data.csv}{\underline{Superhero Power Analytics Dataset}}. ARTEMIS-DA groups superheroes based on their power attributes, leveraging dimensionality reduction and visualization techniques. This task highlights ARTEMIS-DA’s versatility in unsupervised learning, particularly in applications such as customer segmentation, anomaly detection, and clustering-based insights.

    \item \textbf{Time Series Forecasting on Stock Exchange Data:} The middle-left sub-figure demonstrates ARTEMIS-DA’s strength in time-series forecasting. Utilizing an advanced Long Short-Term Memory (LSTM) model, ARTEMIS-DA predicts trends and fluctuations in stock prices using historical data from the \href{https://www.kaggle.com/datasets/mattiuzc/stock-exchange-data?select=indexData.csv}{\underline{Stock Exchange Data}} dataset. This capability highlights ARTEMIS-DA’s applicability to predictive financial analysis, such as stock market forecasting, economic modeling, and investment strategy development.

    \item \textbf{Text Classification of BBC Articles:} The middle-right sub-figures showcase ARTEMIS-DA’s application to natural language processing (NLP), specifically in text classification. Using the \href{https://www.kaggle.com/datasets/alfathterry/bbc-full-text-document-classification?select=bbc_data.csv}{\underline{BBC Document Classification}} dataset, ARTEMIS-DA successfully categorizes news articles by topic. This task underscores ARTEMIS-DA’s ability to process unstructured text data, making it an effective tool for applications like content categorization, sentiment analysis, and automated tagging.

    \item \textbf{Sentiment Analysis on Disneyland Reviews:} The bottom-left sub-figure illustrates ARTEMIS-DA’s ability to perform sentiment analysis on text reviews using the \href{https://www.kaggle.com/datasets/arushchillar/disneyland-reviews?select=DisneylandReviews.csv}{\underline{Disneyland Reviews}} dataset. ARTEMIS-DA effectively identifies the Disneyland location with the highest percentage of positive reviews, showcasing its utility in analyzing customer or consumer sentiment.

    \item \textbf{Regression for Housing Prices:} The bottom-right sub-figures demonstrate ARTEMIS-DA’s regression modeling capabilities using the \href{https://www.kaggle.com/datasets/sukhmandeepsinghbrar/housing-price-dataset?select=Housing.csv}{\underline{Housing Price Dataset}}. ARTEMIS-DA applies feature selection via heatmaps and builds a regression model that achieves an $R^2$ score of 0.78, highlighting its ability to uncover relationships between variables and predict continuous outcomes.
\end{itemize}

These examples illustrate ARTEMIS-DA’s adaptability in addressing a broad spectrum of predictive modeling tasks, from binary classification and NLP to time-series forecasting, clustering, sentiment analysis, and regression. ARTEMIS-DA’s robust performance across different domains makes it an ideal solution for data-driven applications in diverse industries, offering powerful tools for data exploration and predictive analytics, machine learning.

\section{Conclusion}

This paper presents the \textbf{Advanced Reasoning and Transformation Engine for Multi-Step Insight Synthesis in Data Analytics (ARTEMIS-DA)}, a groundbreaking framework that extends the analytical capabilities of Large Language Models (LLMs) to handle complex, multi-step, data-driven queries with minimal user intervention. By integrating the \textit{\textbf{Planner}}, \textit{\textbf{Coder}}, and \textit{\textbf{Grapher}} components, ARTEMIS-DA seamlessly combines high-level reasoning, real-time code generation, and visual analysis to orchestrate sophisticated analytics workflows encompassing data transformation, predictive modeling, and visualization.

Our evaluations highlight ARTEMIS-DA’s state-of-the-art performance on benchmarks such as TabFact\cite{2019TabFactA}, WikiTableQuestions\cite{pasupat-liang-2015-compositional}, and FeTaQA\cite{10.1162/tacl_a_00446}, demonstrating its effectiveness in managing nuanced, multi-step analytical tasks. By decomposing natural language queries into logical tasks, executing precise code for each step and performing visual analysis of generated graphs, the framework bridges the gap between intuitive user interactions and advanced computational execution.

ARTEMIS-DA empowers both technical and non-technical users by simplifying precise analysis of complex datasets. Future directions include enhancing the framework’s adaptability to broader tasks, improving its computational efficiency, and exploring its application in domains such as software engineering, where rapid and accurate multi-step analysis is essential.

\typeout{}
{
    \small
    \bibliographystyle{ieeenat_fullname}
    \bibliography{main}
}

\end{document}